# ИНФОРМАЦИОННАЯ ТЕХНОЛОГИЯ РАСПОЗНАВАНИЯ РУКОПИСНЫХ МАТЕМАТИЧЕСКИХ ВЫРАЖЕНИЙ В РЕЖИМЕ РЕАЛЬНОГО ВРЕМЕНИ НА ОСНОВЕ НЕЧЕТКИХ НЕЙРОННЫХ СЕТЕЙ

## Эдрис Надеран

***Аннотация:*** *в статье рассматривается разработанная информационная технология распознавания рукописных математических выражений, вводимых в ЭВМ в режиме реального времени, в основе которой лежат предложенные подходы к распознаванию рукописных символов и структурному анализу математических выражений.*

***Ключевые слова:*** *распознавание рукописных математических выражений, нечеткая логика, нечеткие нейронные сети, структурный анализ.*

***ACM Classification Keywords:*** *I.5.2 Computing Methodologies - Pattern Recognition - Design Methodology - Pattern analysis. G.4 Mathematics of Computing – Mathematical Software - Algorithm design and analysis. I.5.1 Computing Methodologies - Pattern Recognition - Models - Structural.*

**Вступление**

Математические выражения составляют основную часть в большинстве научных и технических дисциплин, однако на сегодняшний день ввод математических выражений в ЭВМ является сложным и неудобным для пользователя ПК, поскольку осуществляется с помощью традиционных устройств ввода, таких как клавиатура и мышь, к тому же отнимает большое количество времени.

Активное развитие планшетных ПК, наблюдаемое сегодня, приводит к необходимости ввода данных в ЭВМ без использования клавиатуры. По прогнозам компании DisplaySearch к 2016 году объем мировых продаж планшетные ПК превысит объем мировых продаж ноутбуков (Рис. 1). Возможность взаимодействия пользователя с ПК с помощью сенсорной функциональности станет безусловно самой естественной, удобной и быстрой альтернативой для ввода математических выражений в ЭВМ.

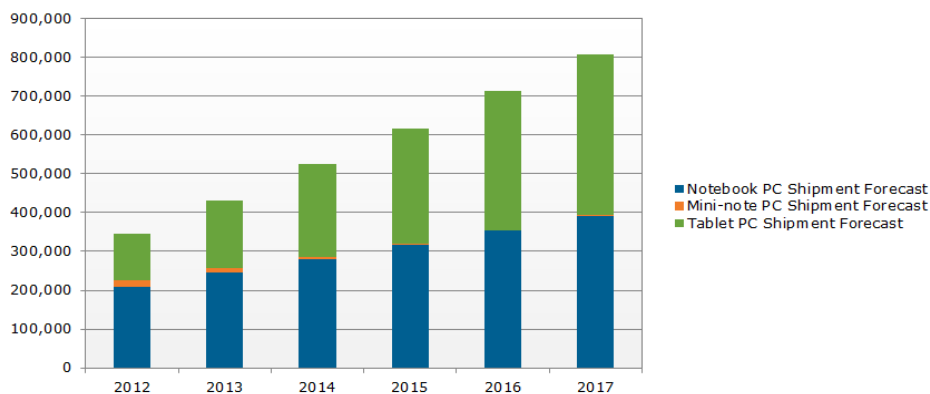

**Рис. 1** Прогноз мировых поставок мобильных ПК (источник NPD DisplaySearch)

Целью данной статьи является рассмотрение разработанной информационной технологии, обеспечивающей эффективное решение задачи распознавания рукописных математических выражений, вводимых в ЭВМ в режиме реального времени.

**Описание и ключевые особенности информационной технологии**

Рассмотрим общую схему работы информационной технологии распознавания рукописных математических выражений, вводимых в режиме реального времени. Процесс распознавания математических выражений состоит из двух основных этапов: распознавания рукописных символов и структурного анализа.

При распознавании рукописных символов в режиме реального времени можно проследить траекторию написания символа и определить точку касания пера и точку отрыва пера при написании. Для получения информативных признаков, характеризующих символы математического выражения, использован предложенный в [Надеран Э., Зайченко Ю.П., 2012] подход, основанный на использовании точек ломаной аппроксимирующей кривую символа. Опытным путем был подобран оптимальный коэффициент точности сглаживания, при котором получено оптимальное соотношение между количеством оставляемых алгоритмом точек и максимальным сохранением очертания символа. В качестве классификатора используется модифицированная ННС NEFCLASS, отличительной особенностью которой является возможность формирования логического вывода на основе аппарата нечеткой логики, что позволяет заложить знания эксперта в систему, при этом параметры функций принадлежности настраиваются с использованием алгоритмов обучения нейронных сетей. Для повышения качества распознавания символов применяется генетический алгоритм обучения параметров функции принадлежности на этапе первичного обучения системы и алгоритм сопряженных градиентов на этапе дообучения системы для улучшения временных показателей [Надеран Э., Зайченко Ю.П., 2012].

Этап структурного анализа позволяет определить пространственные отношения между составляющими математического выражения и включает в себя следующие этапы: размещение, реконструкция и группировка символов [Надеран Э., Зайченко Ю.П., 2013]. Символы разделены на группы, в зависимости от допустимых и обязательных позиций размещения других символов относительно них. На этапе размещения определяется наилучшая позиция размещения для символа по формуле:

$$NP = P * k$$

где P – процент попадания символа

k – коэффициент позиции, принимающий значения:

    0 - для недопустимых позиций;

    1 - для допустимых позиций;

    1,5 - для обязательных позиций.

На этапе реконструкции символов применяется динамическая база эвристических правил, основанная на знаниях о порядке записи и пространственных отношениях между символами, и позволяющая проводить реконструкцию символов и коррекцию неправильно распознанных символов в последовательности путем нахождения ее семантического значения. На этапе группировки некоторые символы позволяют сгруппировать несколько отдельных символов в одну группу. К таким символам относятся: различного вида скобки, математические аббревиатуры, дробная черта, сумма $\Sigma$, произведение$\Pi$, интеграл, арифметический корень, точка и запятая. Написанное пользователем математическое выражение просматривается после каждого внесенного изменения, что дает возможность записывать составляющие математическог о выражения в любом порядке, а также вносить изменения в уже написанное выражение.

Основные требования, которым должна соответствовать информационная технология:

1. Гибкость, способность к адаптации и дальнейшему развитию.
2. Надежность.
3. Минимальное время отклика на запрос.
4. Возможность адаптации к различным почеркам.
5. Возможность работы независимо от платформы.
6. Возможность использования результата распознавания в популярных текстовых редакторах.
7. Удобство и простота интерфейса.
8. Эффективность.

Информационная технология распознавания рукописных математических выражений состоит из следующих компонентов (Рис.2):

1. Event Processing, который обеспечивает многозадачность и синхронную обработку событий. К событиям относятся: распознавание символов, структурный анализ математического выражения, удаление символов, обучение нейронной сети.
2. Training, который включает в себя модуль GA Training, реализующий генетический алгоритм обучения, и модуль CGA Training, реализующий алгоритм сопряженных градиентов обучения нейронной сети.
3. Recognition, который включает в себя модуль Classifier NEFCLASS, реализующий нечеткий классификатор NEFCLASS и модуль Features Processing, вычисляющий значения информативных признаков.
4. Structural Analysis, который отвечает за решение задачи структурирования последовательности введенных пользователем символов и состоящий из модуля Expression Analyzer, отвечающего за подготовку входных и выходных данных, модуля Symbol Arrangement, реализующего этап размещения символов, модуля Reconstruction, отвечающего за реконструкцию символов, и модуля Grouping, обеспечивающего этап группировки символов.

Модуль Database Layer обеспечивает взаимодействие системы с базой данных. В базе данных хранятся параметры обученного с помощью генетического алгоритма обучения нечеткого классификатора NEFCLASS, необходимые для этапа распознавания рукописных символов, написанных с помощью одного штриха. Таким образом, в базе данных хранятся параметры функции принадлежности Гаусса, количество нейронов входного и выходного слоя, а также база нечетких правил. Для проведения структурного анализа в базе данных хранится база эвристических правил, а также для каждого символа хранится информация о допустимых, обязательных и недопустимых позициях расположения других символов относительно них.

При коррекции пользователем результата распознавания символов будет дообучаться нейронная сеть и обновляться параметры классификатора в базе данных, а также при коррекции пользователем результатов структурного анализа будут обновляться данные эвристической базы правил для данного пользователя. Таким образом, предлагаемое решение позволяет персонализировать систему распознавания рукописных математических выражений, повысить эффективность ее работы и подстроить систему под почерк пользователя.

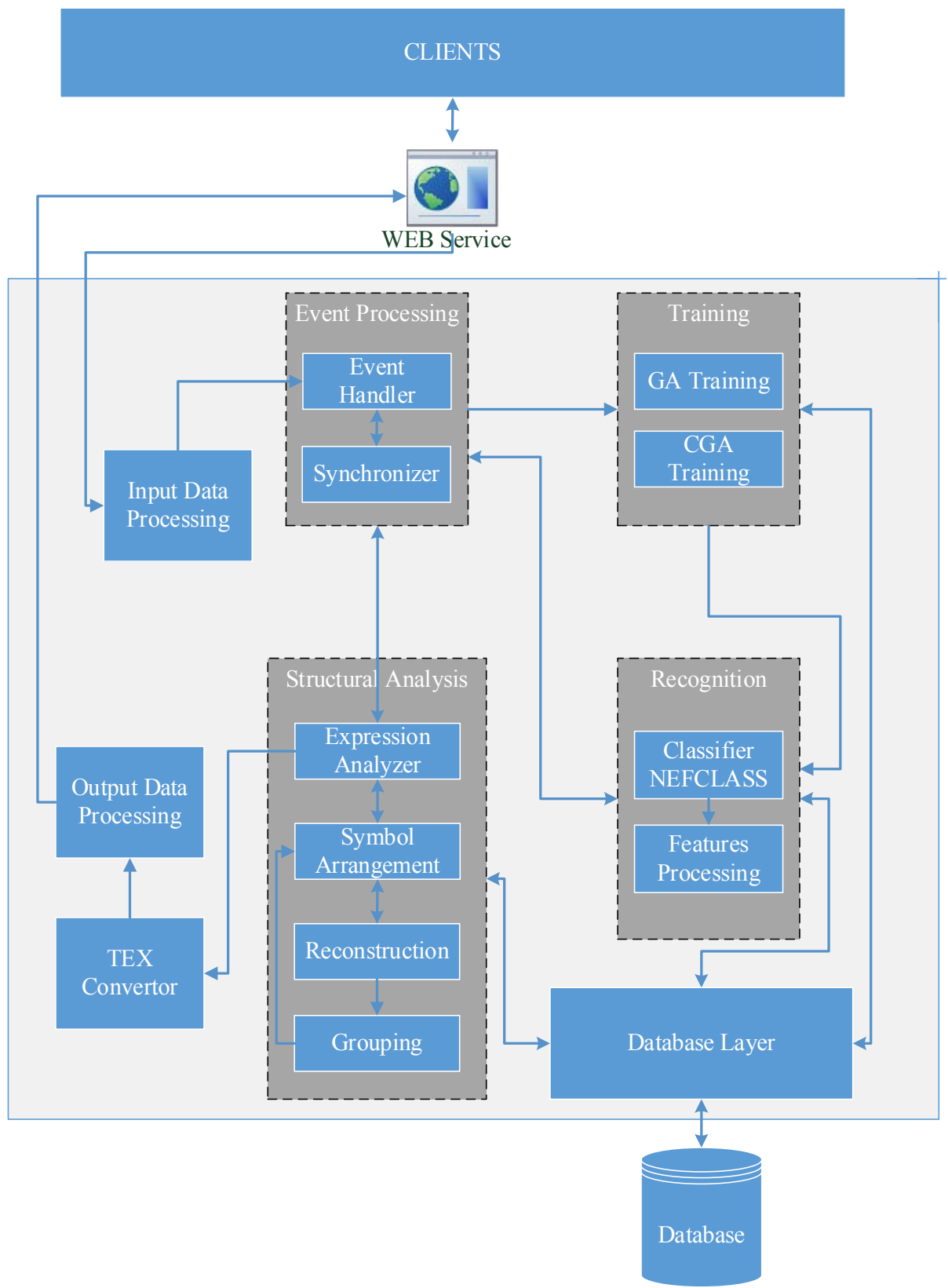

**Рис.2** Структурная схема

**Экспериментальные исследования и результаты тестирования**

Программная часть системы реализована на языке C# на базе платформы Microsoft .NET. На Рис.3 представлена диаграмма классов ИТ. Преимущество использования платформы .NET Framework заключается в возможности реализации исполняемого кода независимого от аппаратной части, поскольку исходный код переводится компилятором в промежуточный байт-код Common Intermediate Language (CIL) и затем код исполняется виртуальной машиной Common Language Runtime (CLR), которая с помощью встроенного в неё JIT-компилятор преобразует промежуточный байт-код в машинные коды нужного процессора. Кроме того, CLR заботится о базовой безопасности, управлении памятью и системе исключений. Необходимо отметить, что использование современной технологии динамической компиляции позволяет достигнуть высокого уровня быстродействия.

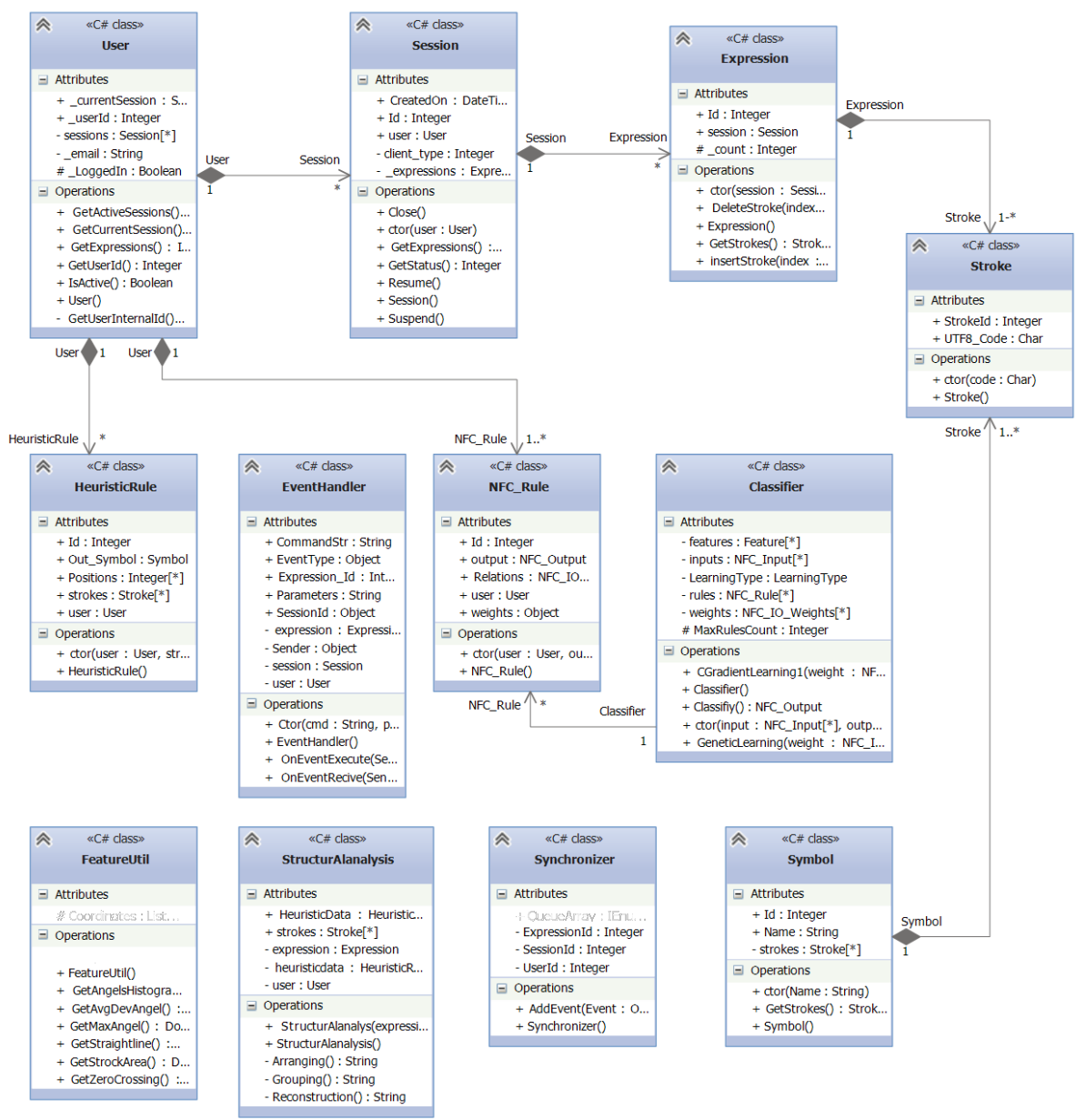

**Рис.3** Диаграмма классов ИТ

Для проверки эффективности работы, предложенной информационной технологии были проведены экспериментальные исследования, позволяющие оценить качество распознавания рукописных математических выражений. Для обучения нечеткого классификатора NEFCLASS была использована выборка, состоящая из 260 рукописных математических выражений. Для генерации базы правил каждому образцу ставился в соответствие подходящий рукописный символ. После чего нейронная сеть была обучена с помощью генетического алгоритма обучения. Таким образом, были найдены значения весовых коэффициентов необходимые для инициализации нейронной сети.

Тестирование проводилось путем рукописного ввода математических выражений в режиме реального времени в планшетный ПК с помощью стилуса. Общая выборка для тестирования составила 140 математических выражений. Среднее время распознавания каждого символа составило 150 миллисекунд на компьютере со следующими характеристиками: CPU Intel Core 2 Quad с частотой 2.4 ГГц, RAM 4 Гб. Результаты тестирования отражены на Рис.4 и отражают качество распознавания и структурирования математических выражений.

| Распознавание отрезков | Реконструкция символов (для правильно распознанных отрезков) | Структурный анализ математических выражений |
|---|---|---|
| 91,54% | 95,32% | 71,29% |

**Рис.4** Результаты тестирования

Поле "Распознавание отрезков" показывает процент правильно классифицированных символов, написанных без отрыва пера с помощью одного штриха, и позволяет оценить эффективность распознавания. Поле "Реконструкция символов" отображает процент верно реконструированных символов из распознанных правильно отрезков. Поле "Структурный анализ математических выражений" показывает процент правильно структурированных выражений.

**Заключение**

В статье рассмотрена разработанная информационная технология распознавания рукописных математических выражений, вводимых в ЭВМ в режиме реального времени, в основе которой лежат предложенные подходы к распознаванию рукописных символов и структурному анализу математических выражений. Описаны основные требования, которым должна соответствовать информационная технология и рассмотрены компонентов из которых она состоит. Описаны проведенные экспериментальные исследования рассмотренной информационной технологии. Среднее время распознавания каждого символа составило 150 миллисекунд, что удовлетворяет требованиям к использованию системы в режиме реального времени. Точность распознавания составила 91,54%, точность структурного анализа составила 71,29%.

**Благодарности**

**Информация про автора**

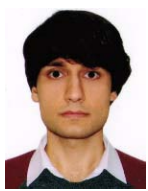

*Надеран Эдрис* –аспирант Национального технического университета Украины "КПИ".

*Основные сферы научных исследований автора*: применение нечеткого классификатора NEFCLASS к задаче распознавания рукописных математических выражений.

# Online Handwritten Mathematical Expressions Recognition System Using Fuzzy Neural Network

## Edris Naderan

**Abstract:** The paper is devoted to the development of the new online handwritten mathematical expressions recognition system. The paper presents the recognition method to the handwritten symbols using fuzzy neural network NEFCLASS as a means for classification.

**Keywords:** online handwriting recognition, mathematical expressions, fuzzy logic, fuzzy neural networks, structural analysis, artificial neural network, genetic algorithm, conjugate gradient algorithm.